\documentclass[letterpaper, 10 pt, conference]{ieeeconf}  % Comment this line out if you need a4paper

\IEEEoverridecommandlockouts                              % This command is only needed if 
\usepackage{graphics} % for pdf, bitmapped graphics files
\usepackage{epsfig} % for postscript graphics files
\usepackage{mathptmx} % assumes new font selection scheme installed
\usepackage{times} % assumes new font selection scheme installed
\usepackage{amsmath} % assumes amsmath package installed
\usepackage{amssymb}  % assumes amsmath package installed
\usepackage{xcolor}
\usepackage{listings}
\usepackage{multirow}
\usepackage{booktabs}
\usepackage{amsfonts}
\usepackage{caption}
\usepackage{lipsum} % For generating dummy text
\usepackage{tcolorbox}
\let\labelindent\relax
\usepackage{enumitem}  % Load the enumitem package
\setlist[itemize]{topsep=0pt, left=0pt} 
\usepackage{newtxmath}
\newcommand{\modelname}{ORION}
\newcommand{\taskname}{ZIPON}

\title{\LARGE \bf
Think, Act, and Ask: Open-World Interactive Personalized Robot Navigation
}

\author{Yinpei Dai$^\dagger$, Run Peng$^\dagger$, Sikai Li$^\dagger$, Joyce Chai$^\dagger$% <-this % stops a space
\thanks{$^\dagger$Computer Science and Engineering Division, University of Michigan. Emails: \texttt{\{daiyp,chaijy\}@umich.edu}.}%
\thanks{This work is supported by Amazon Consumer Robotics, NSF IIS-1949634, NSF SES-2128623, and has benefited from the Microsoft Accelerate Foundation Models Research (AFMR) grant program.}% <-this % stops a space
}

\begin{document}

\maketitle
\thispagestyle{empty}
\pagestyle{empty}

\begin{abstract}
Zero-Shot Object Navigation (ZSON) enables agents to navigate towards open-vocabulary objects in unknown environments. 
The existing works of ZSON mainly focus on following individual instructions to find generic object classes, neglecting the utilization of natural language interaction and the complexities of identifying user-specific objects.
To address these limitations, we introduce Zero-shot Interactive Personalized Object Navigation (\taskname), where robots need to navigate to personalized goal objects while engaging in conversations with users. To solve \taskname, we propose a new framework termed Open-woRld Interactive persOnalized Navigation (\modelname) \footnote{Code available at https://github.com/sled-group/navchat}, which uses Large Language Models (LLMs) to make sequential decisions to manipulate different modules for perception, navigation and communication. Experimental results show that the performance of interactive agents that can leverage user feedback exhibits significant improvement.
However, obtaining a good balance between task completion and the efficiency of navigation and interaction remains challenging for all methods. We further provide more findings on the impact of diverse user feedback forms on the agents' performance.

\end{abstract}
% \jycc{I de-emphasize zero-shot as most recent works all zero-shot. The key here is interaction, personalization. Open world somewhat implies zero-shot.}

\section{INTRODUCTION}
Recent years have seen an increasing amount of work on Zero-Shot Object Navigation (ZSON) where an embodied agent is tasked to navigate to open-vocabulary goal objects in an unseen environment \cite{majumdar2022zson, zhou2023esc}. This task is crucial for developing robots that can work seamlessly alongside end-users and execute open-world daily tasks through natural language communication. A common practice in ZSON is to utilize pre-trained vision-language models (VLMs) out-of-the-box to ground natural language to visual observations, thus enabling the robots to handle unseen object goals without additional training \cite{cow,pasture,nlmap,vlmap}. 
%While promising zero-shot performance is achieved, these previous works mainly focus on following single user instructions to find generic object classes, leading to several limitations.
Despite recent advances, several limitations remain which hinder the real-world deployment of these agents. 

%This requires two essential characteristics for the deployed agents: (i) good zero-shot abilities on unrestricted language-described tasks and (ii) rapid adaptation to the diverse user language feedback. 
% A common practice in recent work to tackle this problem is to leverage pre-trained large language models (LLMs) \cite{brown2020language} and vision-language models (VLMs) \cite{radford2021learning}, where the LLMs are integrated to decompose the user's commands into atomic actions \cite{saycan} or code snippets \cite{liang2023code} that the system can process and the VLMs are used out-of-the-box to ground the open-vocabulary language description to visual observations \cite{cow}. However, many of the work \cite{pasture, vlmap, zhou2023esc} primarily focused on building autonomous agents to accomplish zero-shot tasks by themselves without considering utilizing user language feedback, which may prohibit safe and rapid prototype in real-world applications.
%while neglecting the vital role of user language feedback for task optimization. 
%Although the performance of many zero-shot tasks like object navigation \cite{majumdar2022zson} and pick-place manipulation \cite{liang2023code} have been substantially enhanced, this paradigm still suffers from two drawbacks. 

\begin{figure}[t]
    \centering
    \includegraphics[width=0.8\linewidth]{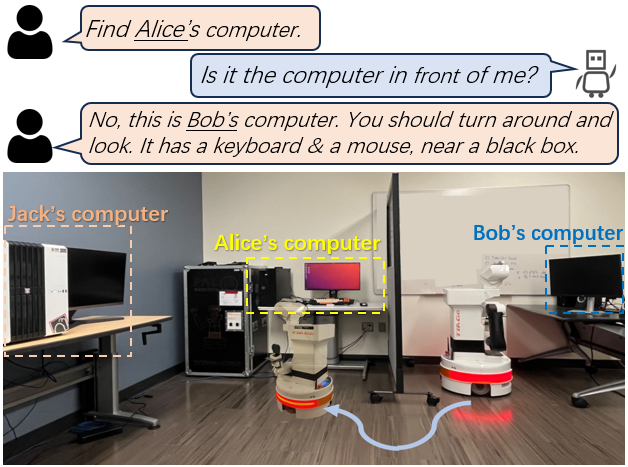}
    \caption{An example of zero-shot interactive personalized navigation. There are three computers in the room never seen by the robot before. The goal is to find \textit{Alice's computer}. The robot starts by finding the wrong object and needs to communicate with the user and leverage the user feedback to locate the personalized goal. 
% \jycc{can you change the figure. the robot feedback: .... You should turn around and look. It has a keyboard \& a mouse, ...}
}
    \label{fig:teaser}
    \vspace{-16pt}
\end{figure}

First, previous works only focus on following individual instructions without considering feedback and interaction. In the realistic setting, instruction-following often involves back-and-forth interaction to reduce uncertainties, correct mistakes, and handle exceptions.   For example, given an instruction \textit{``go to the living room and find the toy airplane"}, if the robot goes to a wrong room, immediate feedback from the user will save the robot's endless search in the wrong room and direct it to the desired location.  Therefore, it is important to build agents that can elicit and leverage language feedback from users during task execution to avoid errors and achieve the goal. Moreover, current navigation tasks are often designed to find any instance of the same object class \cite{majumdar2022zson, habitatchallenge2022, habitatchallenge2023}. However, in the real world, especially in a household, or office setting, objects can often be described by unique and personalized properties that are shared by people with the same background knowledge. For example, as shown in Figure~\ref{fig:teaser}, the robot is instructed to find \textit{``Alice's computer''} (or \textit{``the computer purchased last year"}). Just like humans, the robot needs to acquire and apply such personalized knowledge in communication with users.  Even though current models can correctly identify a `computer' based on their general perception abilities, it remains unclear how to build an agent that can swiftly adapt to a personal environment and meet users' personalized needs.

To this end, we introduce  \textit{Zero-shot Interactive Personalized Object Navigation} (\taskname), an extended version of ZSON. In \taskname, the robot needs to navigate to a sequence of {\em personalized goal objects} (objects described by personal information) in an unseen scene. As shown in Figure \ref{fig:teaser}, the robot can engage in conversations with users and leverage user feedback to identify the object of interest.  Different from previous zero-shot navigation tasks \cite{majumdar2022zson, pasture, vlmap}, our task evaluates agents on two fronts: (i) language interactivity (understanding when and how to converse with users for feedback) and (ii) personalization awareness (distinguishing objects with personalized attributes such as names or product details).
% ; and (iii) online adaptability (the agent should be able to gain information from past navigation experience and consistently improve its performance during the testing). 
% Both simulated users and real users are employed for comprehensive task evaluation.
To the best of our knowledge, this is the first work to study personalized robot navigation in this new setting.

To enable \taskname, we develop a general LLM-based framework for \underline{O}pen-wo\underline{R}ld \underline{I}nteractive pers\underline{O}nalized \underline{N}avigation (\modelname). Specifically, \modelname\ consists of various modules for perception, navigation and communication.  The LLM functions as a sequential decision maker to control the modules in a \textit{think-act-ask} manner: In the \textit{think} step, the LLM reflects the navigation history and reason about the next plans; In the \textit{act} step, the LLM predicts an action to execute a module and the executed message is returned as context input for the next action prediction; In the \textit{ask} step, the LLM generates natural language responses to interact with the user for more information. This framework allows us to incorporate different baselines to conduct extensive studies on  \taskname. Our empirical results have shown that agents that can communicate and leverage diverse user feedback significantly improve their success rates. 
To summarize, the main contributions of the paper are as follows:
\begin{itemize}
    \item We propose a novel benchmark called ZIPON for the zero-shot interactive personalized object navigation problem.
    \item We design a general framework named ORION to perform function calls with different robot utility modules in a think-act-ask process.
    \item We provide insightful findings about how user language feedback influences task performance in ZIPON. 
\end{itemize}

\section{Related Work}
\textbf{Zero-Shot Object Navigation.} 
%Object navigation aims to control a robot to move to a goal object in the environment. Recently, object navigation in zero-shot settings attracts increasing attention. By applying the impressive zero-shot abilities of large pre-trained vision and language models (VLMs) like CLIP[] and LSeg [], any open-vocabulary object that can be described in natural language can be visually grounded to for robot navigation. 
Recently, there has been a growing interest in zero-shot object navigation using VLMs. One line of the approaches is exploration-based, where the agent moves around based on standard exploration algorithms and matches the ego-observations with language descriptions via VLMs \cite{majumdar2022zson, cow, lmnav}. 
% For example, CoW \cite{cow} used the weighted gradients of CLIP \cite{radford2021learning} for object localization in RGB images, LM-Nav \cite{lmnav} used CLIP to select snapshots of possible landmarks along the path.
%ZSON \cite{majumdar2022zson} use CLIP to construct a joint semantic goal embedding space to leverage both object navigation and image navigation datasets. 
Another line is the map-based method, where a spatial semantic map is built with VLM representations to enable natural language indexing \cite{vlmap, jatavallabhula2023conceptfusion, peng2023openscene}. 
% For instance, 
% %NLMap \cite{nlmap} established a querable map enriched with CLIP and ViLD \cite{vild} features for robot planning, 
% VLMap \cite{vlmap} fused LSeg \cite{lseg} features of each image with a standard 3D reconstruction of the physical world, ConceptFusion \cite{jatavallabhula2023conceptfusion} integrated both local and global pixel-aligned features with CLIP to capture long-tailed concepts. 
Our framework integrates both methods for more flexibility and superior navigation performance.

\textbf{Dialogue Agents for Robots.}
%With the off-the-shelf conversing and reasoning  capabilities of large language models, it is not relatively easy to build a robot based on LLMs that understands user intents and schedules all API calls to make actions, and response with users in natural language. 
Dialogue agents enable human-machine conversations through natural language \cite{he2022galaxy,he2022space,he2022unified, wang2022task, si2023spokenwoz}. A large number of earlier works have studied language use in human-robot dialogue~\cite{jia2014icra,scheutz2011aimag, thomason2015ijcai}. Recently, LLMs have been widely used in robots \cite{yang2023llm}.
InnerMonologue \cite{huang2022inner} used an LLM to form an inner monologue style to ask questions. 
PromptCraft \cite{chatgptforrobot} explored prompt engineering practices for robot dialogues with ChatGPT. 
KNOWNO\cite{knowno} measured the uncertainty of LLM planners for agents to ask for help when needed. In contrast, we use LLMs to operate different robot modules and frame it as a sequential decision-making problem.
% to think, act and ask by sequentially calling functions.
%Unlike these work, the LLM in our agent not only can talk and plan, but also can manipulate an additional memory module to memorize the robot errors in the dialogue history, which has the potential for the online continual learning after the robot deployment in new environment. 

\textbf{Leverage Language Feedback.} In human-robot dialogue, robots that can learn from and adapt to language feedback can make more reliable decisions~\cite{she2017acl, she2014sigdial, chai2016aimagazine}.
% and avoid possible catastrophic mistakes. 
Previous works have emphasized real-time robot plan adjustments \cite{sharma2022correcting, cui2023no}. 
% One line of the work focuses on robot plan correction during online execution. \cite{sharma2022correcting} designed a mapping from natural sentences to cost function transformations to accommodate errors.
% or extra constraints. 
%REFLECT \cite{liu2023reflect} used LLMs to summarize robot experiences for failure explanation. 
% \cite{cui2023no} performed online corrections for robotic manipulation via shared autonomy with teleoperation. Another line of the work uses various language hints to help robots. 
Others harness language instructions for assistance \cite{thomason2020vision, pasture}, task learning~\cite{chai2018ijcai}, and human-machine collaboration \cite{HARI, dai2020learning, chai2014hri}. However, no study has comprehensively compared different feedback types in the navigation context.
% CVDN \cite{thomason2020vision} contains movement instructions to guide the navigation. Pasture \cite{pasture} benchmarks a set of uncommon object navigation by describing the goals' appearance and locations. HARI \cite{HARI} proposed to learn to elicit nearby landmarks. However, none of the works thoroughly compared different feedback types on the same navigation task. 

%\textbf{User Simulator.} Real user interaction process can be tedious adnd time-consuming. Therefore, simulated users are often used as a substitute to collect much more abundant data to evaluate dialogue systems in task-oriented settings \cite{budzianowski-etal-2018-multiwoz}. ABUS \cite{keizer2010parameter} presented an agenda-based user simulator that combines handcrafted design and machine learning-based parameter estimation to mimic real user behaviours. NUS \cite{kreyssig-etal-2018-neural} proposed an end-to-end LSTM-based neural user simulator that can learns directly from raw dialogues. JOUST \cite{tseng-etal-2021-transferable} demonstrated the effectiveness of joint learning of both a user simulator and a dialogue agent on domain transfer. But the research of user simulator for robot navigation is still limited. In this paper, we will show how to build a user simulator based on LLMs to facilitate zero-shot object navigation.

\textbf{Personalized Human-Robot Interaction.} Building personalized robots is an active research area in human-robot interaction  \cite{1374720, clabaugh2018robots, hellou2021personalization}. There are many works focused on enhancing personalized experience through non-verbal communications \cite{de2022learning, de2023learning, goetz2003matching} and better interactive service design \cite{lee2012personalization, cakmak2012}.  Personalized dialogue systems also gained increasing interest \cite{zheng2019personalized,schmidt2018survey,mi2021towards,luo2019learning}, where the persona is taken as conditional input to produce more characterized and sociable conversations. To our knowledge, previous work has not investigated interactive personalized navigation tasks.

\section{Interactive Personalized Navigation}
% \jycc{Interactive Persoalized Navigation. I thinks this section needs another re-organization. Here, it should be about (1) the problem definition, (2) framework to solve the problem, (3) the focused types of feedback, and (4) evaluation. you can move Figure 2 here and use it as an example to discuss the problem, solutions, evaluation etc. I feel that Personalized Goal Curation does not belong to this section. This section is really about the conceptual level discussion. Goal curation is how you build the datasets for experiments, which could be the first subsection in the experiment section. }

% We first introduce the task definition then present the open-world interactive personalized navigation framework. Finally, we introduce different types of user feedback we focused on in this work and the evaluation protocol. 

\begin{figure}[t]
    \centering
    \includegraphics[width=\linewidth]{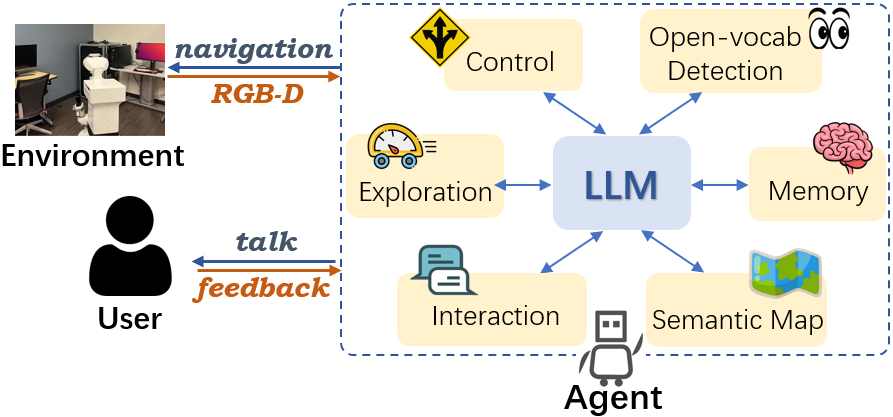}
    \caption{\small The \modelname\ framework architecture. The LLM makes sequential decisions to operate different modules to search, detect and navigate in the environment and talk with the user.}
    \vspace{-15pt}
    \label{fig:framework}
\end{figure}

We first introduce the zero-shot interactive personalized object
Navigation (\taskname) task, then explain the open-world interactive personalized navigation (\modelname) framework.

% Zero-shot multi-goal personalized navigation aims to control robots to navigate to a series of personalized goal objects in unseen scenes by interacting with users in multi-round dialogue, without any additional training. This contrasts to prior zero-shot single-goal navigation task \cite{pasture} and multi-goal non-personalized navigation task \cite{vlmap}, and is closer to the real-world where the user may ask the robot to go to someone's object instead of any object with the same type.

\subsection{The \taskname\ Task}
\taskname\ is a generalized type of zero-shot object navigation \cite{pasture}. Let $\mathcal{E}$ denote the set of all test scenes, and $\mathcal{G}$ denote the set of all personalized goals.
Each goal $g\in G$ contains a tuple $g=(\textit{type}, \textit{name}, \textit{room}, \textit{FB})$, where \textit{type} is the 
class label for the object (e.g., `bed', `chair'), \textit{name} is the personalized expression (e.g. `Alice's bed', `chair bought from Amazon') which is unique for every goal, \textit{room} is the name of the room (e.g., `Alice bedroom', `living room') where $g$ is located in, and \textit{FB} is a dictionary to store all types of user feedback information (See Sec. \ref{sec: feedback}).
A navigation episode $\tau \in \mathcal{T}$ is a tuple $\tau=(e,g,p_{0})$, where $e\in \mathcal{E}$, $g\in \mathcal{G}$, $p_{0}$ is the initial pose of the agent for current $\tau$. The input observations are RGB-D images. Starting from $p_{0}$, the agent needs to find $g$ by taking a sequence of primitive actions, where the action space is  $\mathcal{A}_{task}$=\{\textsc{TurnLeft} 15$^\circ$, \textsc{TurnRight} 15$^\circ$, \textsc{MoveForward} 0.25m, \textsc{Talk}\}. The first three are navigation actions, and the last one is a communicative action with the dialogue content to be generated by the agent.
During the evaluation for $\tau$, the agent can either move in the environment with navigation actions or interact with the user with \textsc{Talk} for more information. Whenever the agent believes it has reached the goal, it must issue \textsc{Talk} to stop moving and confirm with the user. 
The $\tau$ is terminated when the robot successfully finds the $g$ or a maximum number $I_{\max}$ of interaction attempts (i.e., the total number of \textsc{Talk} actions) is attained. If the agent is within $c$ meters of $g$ and meets visibility criteria, the $\tau$ is successful.
% \jycc{I found definition of $\tau$ is not clear. It's not a sequence, it cannot stop. It's more like a problem definition, defining a task. the solution is a sequence of primitive actions. }
% \dyp{I follow the \cite{pasture} }

\subsection{Proposed \modelname\ Framework}
We propose \modelname\ to solve \taskname. As shown in Fig. \ref{fig:framework}, this framework comprises six modules:  a control module to perform navigation movements, a semantic map module for natural language indexing, an open-vocabulary detection module to detect any objects
with language descriptions, an exploration module to search the room, a memory module to store crucial information from user feedback, and an interaction module to talk. Central to these, the Large Language Model (LLM) serves as the primary controller, making sequential actions based on its strong reasoning abilities. 
% Fig. illustrates the overall architecture. 
% so that the robot can detect open-set object perform navigation in the environment and interact with the user. 

%We have a trajectory planner module to  predict suitable navigation path given the a 2D destination point in the top-down occupancy grid map, a semantic map module to match goal area of the target object in natural language, a memory module to store the useful information of objects during the 

\subsubsection{LLM Action Space}
To make \modelname\ more general, we design a unified action space for LLM to manipulate all modules. 
% The LLM is responsible to predict actions by generating text strings
An \textit{LLM action} is defined as a string indicating a python function, e.g., \texttt{\small FuncName(param1=value1, $\dots$)}. The total LLM action space, denoted as $\mathcal{A}_{LLM}$, is a collection of all executable functions for LLM, extending beyond $\mathcal{A}_{task}$ to encompass more high-level abilities.
On receiving the user utterance, the LLM will generate the next action to execute a specific module and take the returned function message as new input to predict the next action until the \textsc{Talk} action is chosen; then the agent will communicate with the user, asking for confirmation or more information.
To best utilize the powerful reasoning abilities of LLMs, we propose a \textit{think-act-ask} mechanism to prompt the LLM to generate actions. Specifically, given context, the LLM yields a JSON-format string like \text{\small \{``Thought":$\cdots$, ``Action":$\cdots$\}} for the next \textit{action}. This string will be parsed, and the \text{\small ``Action"} part is used to operate modules for navigation or interaction. 
Below is an example of the LLM context, where text color differentiates input (black) and output (red).
\lstset{
  backgroundcolor=\color{gray!20},
  basicstyle=\scriptsize\mdseries\ttfamily,
  breaklines=true,
  moredelim=**[is][\color{red}]{@}{@},
  moredelim=**[is][\color{cyan!20}\bfseries]{|}{|}
}
\begin{lstlisting}
... # definition of the task
... # definition of all LLM actions

1 Here is an example to perform sequential actions:
2 User Utterance: Find Alice's computer.
3 @{"Thought":"The user wants Alice's computer, I should first search the memory to see if I found it before",@
4  @"Action":{"name":"retrieve_memory", 
             "args":{"obj_str":"Alice's computer"}}}@
5 Function Return: Found 0 items in memory: []
... # internal process of the LLM
6 @{"Thought":"I found a possible computer in the room, it might be the correct one, I shall ask the user",@
7  @"Action":{"name":"talk", 
        "args":{"content":"Is this Alice's computer?"}}}@
8 Robot Response: Is this Alice's computer?
9 User Utterance: No, it's Bob's computer. Keep searching
... 
\end{lstlisting}
The initial two lines expound the \taskname\ task and LLM actions (omitted for brevity). Line 1 begins the illustration of \textit{think-act-ask} process. Line 2 is the first user input.
% and Line 5-6 is the LLM action output that includes the thoughts about how to make proper decisions and the complete function calling string. 
Lines 3-4 show the LLM's subsequent thought process and resultant action.
Line 5 is the executed message from the last action. 
% Line 8 indicates the internal process of the LLM to generate a series of actions to perceive and navigate in the environment before talking to the user, which we omit here for space.
The line following line 5, omitted for space, signifies the internal procedure for generating sequential actions before user interaction, usually with multiple rounds.
Line 6-7 shows the LLM action output for communication when the LLM thinks it's time to talk. The robot will then interact with the user and continue to take action based on the new user input (line 9). 
% Lines 6-7 display the LLM action output for communication, initiated when the LLM determines it's time to inquire.  The robot then engages with the user and continue \textit{think-act-ask} based on subsequent user feedback (line 9).
Following this design, the LLM can manipulate all modules seamlessly after the user issues a goal and decide when and what to talk with the user.
% whether to elicit more information or attempt to confirm the found object with the user to finish the current episode.

\subsubsection{Operated Modules} In this section, we delve into the utilities and LLM actions associated with each module.
% Appendix \ref{sec: appx-llm-functions} gives a more detailed definition of actions.

\textbf{Control Module.} It contains two low-level navigation actions, \texttt{\small Move(num)} and \texttt{\small  Rotate(num)}, that cover the primitive navigation actions in $\mathcal{A}_{task}$. Concretely, \texttt{\small  Move(1)} is \textsc{MoveForward} 0.25m, \texttt{\small  Rotate($\pm$1)} denotes \textsc{TurnRight} 15$^\circ$ and \textsc{TurnLeft} 15$^\circ$, respectively.
In addition, it has two high-level actions: \texttt{\small  goto\_point(point)}, that sequences low-level navigation actions to navigate to a point on a 2D occupancy map, and \texttt{\small  goto\_object(obj\_id)} that directly navigates to an object with the given id.
% acquired from other modules.

\textbf{Semantic Map.} 
% Previous works \cite{vlmap,jatavallabhula2023conceptfusion} have shown that building a spatial map with neural features for natural language query can substantially improve the zero-shot navigation performance. 
Following \cite{vlmap}, we collect RGB-D images to build vision-language maps by reconstructing the point cloud and fusing the LSeg \cite{lseg} features from each point to yield a top-down neural map denoted as $M_{sem}\in \mathbb{R}^{L\times W\times C}$. Here, $L$ and $W$ are the map length and width, respectively. $C$ is the CLIP \cite{openclip} feature dimension. 
The LLM action for this module, \texttt{\small  retrieve\_map(obj\_str)}, utilizes the CLIP text encoder to obtain the embedding $t\in \mathbb{R}^{C}$ for an object description string. It then identifies a similarity matching area $A_{sem}\in \mathbb{R}^{L\times W}$ in the map and extracts suitable contours. These contours are returned as a list of tuples (obj\_id, distance, angle), indicating the assigned ID,  distance and angle of the contour's center relative to the agent's pose.
% The LLM action for this module, \texttt{\small  retrieve\_map(obj\_str)}, employs the CLIP text encoder to obtain the embedding $t\in \mathbb{R}^{C}$ of any object description string and acquire a similarity matching area $A_{sem}\in \mathbb{R}^{L\times W}$ in the map, then finds suitable contours in $A_{sem}$ and returns them as a list of tuples (obj\_id, distance, angle) indicating an assigned id, relative distance and angle of the contour mass centre with respect to the agent pose.

\textbf{Open-vocabulary Detection.} This module is to detect any object from an RGB image using its language description. 
% The returned detected results can increase more situated awareness for LLMs. 
We use the grounded-SAM \cite{liu2023grounding}  
due to its good performance,  
% on many open-set vision benchmarks. However, 
but any other detection models 
% that localize objects in the image to give natural descriptions 
can be used here. Once detected, 
the segmented pixels are transformed into 3D space and projected to the 2D occupancy map to acquire the detected area $A_{det}\in\mathbb{R}^{L\times W}$. The module has two actions: (i) \texttt{\small  detect\_object(obj\_str)} that returns a list of detected objects written as $(\text{obj\_id, distance, angle, detection\_score})$ tuple, and (ii) \texttt{\small  double\_check(obj\_id)} which repositions to a closer viewpoint of an object and detect again.

\begin{figure*}[ht]
    \centering
    \includegraphics[width=\textwidth]{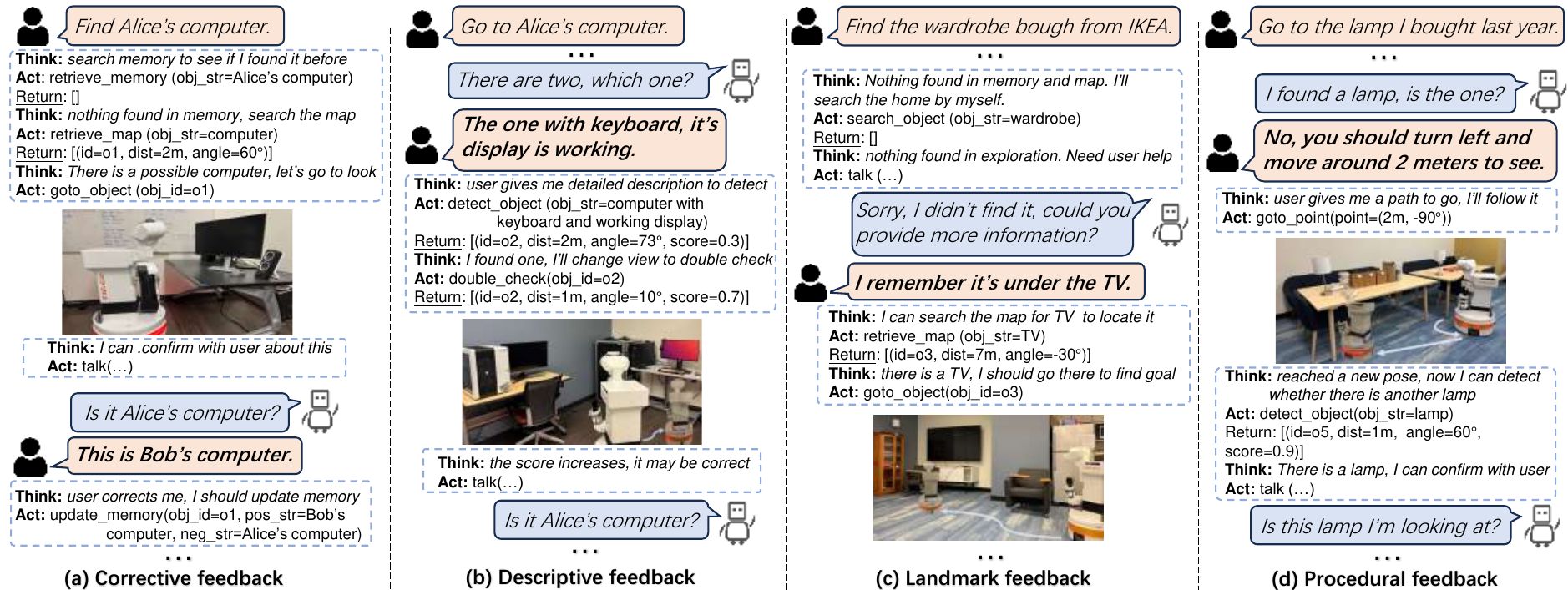}
    \caption{The dialogue examples for different language feedback types. Corrective feedback indicates errors in the robot's object identification. Descriptive feedback details the goal's appearance and status. Landmark feedback points out salient nearby objects. Procedural feedback suggests a rough navigation path to the goal. 
    The content within the dashed box represents the internal \textit{think-act-ask} processes the LLM undergoes. The content in the coloured boxes denotes the interactions between the user and the robot. Ellipsis marks indicate the omission of intermediate human/robot dialogue contents and LLM internal process for space brevity. 
    % In our \modelname\, the LLM makes sequential actions to operate different modules according to the user utterance and the returned function messages.
    }
    \label{fig: large_dialog}
    \vspace{-15pt}
\end{figure*}

% Specifically, we first collect RGB-D images in the scene by standard exploration (See next module) and reconstruct the global point cloud from the depth image. Each point corresponds to a pixel in the RGB image, where the LSeg model \cite{lseg} is used to extract the pixel-wise feature. After fusing the features of all points from images, the point cloud is voxelized and projected into a top-down neural map, 
% which denotes as $M_{sem}\in \mathbb{R}^{L\times W\times C}$. $L$, $W$ are the map length and map width respectively, $C$ is the CLIP feature dimension.
% Since the semantic spaces of LSeg and CLIP are aligned, we can perform language indexing by using CLIP text encoder to get text embedding $t\in \mathbb{R}^{C}$ and acquire the similarity matching area $A_{map}\in \mathbb{R}^{L\times W}$ in the map.

\textbf{Exploration Module.}  
% This module facilitates agent exploration in unseen scenes using frontier-based exploration (FBE) \cite{yamauchi1997frontier}, a heuristic method equalling the efficacy of learning-based approaches in zero-shot navigation \cite{cow}. 
This module uses frontier-based exploration (FBE) \cite{yamauchi1997frontier} to help the agent search unseen scenes.
% that matches the efficacy of learning-based methods in zero-shot navigation \cite{cow}. 
Following \cite{cow}, the agent starts with a 360$^\circ$ spin and builds a 2D occupancy grid map for exploration.
% In the beginning of $\tau$, the agent will spin around 360$^\circ$ to visualize the surroundings, then use depth images to classify the space into a 2D occupancy grid map, where the grid types are free, occupancy or unknown. The boundaries between free and unknown space are called frontiers, whose centres guide the robot's navigation. 
% and stop when there is no unexplored grid on the map.
The LLM action for this module is \texttt{\small  search\_object(obj\_str)}, where FBE determines the next frontier points for the control module to take \texttt{\small  goto\_point} and the detection module to perceive on-the-fly with \texttt{\small  detect\_object}. 
Upon detecting objects, this module returns the detection message. If invoked again, it continues to explore the room until FBE stops.
% there are no unexplored grids on the map.
% performed to generate the next points for the control module to take \texttt{\small  goto\_point}, and the detection module will be used on the fly to perceive possible goals with \texttt{\small  detect\_object}.

%Concretely, it utilizes the grounding DINO [], a state-of-the-art object detection model to predict the possible bounding boxes in the image, and then uses the powerful Segment-Anything Model (SAM) [] to get the fine-grained segmentation of the target object. With the segmented results and the depth image, we can easily construct the predicted area in the top-down map for the object.
%However, one drawback of grounded-SAM is the high time latency. To enable real-time detection, we use the CLIP model to compute the similarity score of the input natural description and the RGB image, only those images higher than a threshold will be fed to grounded-SAM for processing. 

\textbf{Memory.} 
This module stores crucial user information with two neural maps, $M_{pos}$ and $M_{neg}$, both of which have the same size as $M_{sem}$ and hold CLIP features. $M_{pos}$ saves user-affirmed information, while $M_{neg}$ records user denials.
Both maps begin with zero initialization and are updated through the action \texttt{\small  update\_memory(obj\_id, pos\_str,neg\_str)}.
For example, if the user confirms \textit{``yes, it's Alice's desk"}, then the LLM sets \textit{``Alice's desk"} as the positive string and adds its CLIP text feature into the object area associated with the given object id in $M_{pos}$. Conversely, a denial like \textit{``no, it's not cabinet"} leads to the CLIP feature of \textit{"cabinet"} being stored in $M_{neg}$ as the negative string. Object areas derive from either $A_{sem}$ or $A_{det}$. Another action, \texttt{\small  retrieve\_memory(obj\_str)}, matches the CLIP text embedding of the input object description string with features in two maps, then returns retrieved objects as (obj\_id, distance, angle) tuples from $M_{pos}$ while avoiding those in $M_{neg}$.

\textbf{Interaction.} This module is for the communicative interface between the robot and the user.
This work uses a textual interface where users interact with the robot by typing texts. 
The LLM action is \texttt{\small  talk(content)}, which is the same \textsc{Talk} in $\mathcal{A}_{task}$ with the content to be generated by the LLM.

\section{Experiments}

In this section, we begin by introducing the experimental design for \taskname\  in Sec. \ref{sec: exp_design}. Then, we elaborate on the experiments conducted in simulated environments in Sec. \ref{sec: exp_sim} and demonstrate our method on real robots in Sec. \ref{sec: realrobot}.

\subsection{Experimental Design}
\label{sec: exp_design}
Natural language interaction is the core feature of our \taskname\  task. When the robot makes mistakes or asks for help, the user can guide the robot to find the goal through diverse forms of language feedback.
Therefore, we conduct experiments to examine the influence of different types of user feedback on \modelname\  and baseline alternatives.

\textbf{User Feedback Types.}
\label{sec: feedback}
Four common types of user language feedback are used in our experiments:
1) \textit{corrective feedback}, which tells the robot what object it has actually reached if that’s not the correct goal;
2) \textit{descriptive feedback}, which provides more details about the goal object's appearance, status, functions, etc; 
3) \textit{landmark feedback}, which gives the object landmarks like other common and salient objects near the current goal object; 
and 4) \textit{procedural feedback}, which offers language-described approximate routes to instruct the robot to approach the goal from the current pose step by step. We hope through controlling the form of user feedback during the interaction, some interesting findings can be generated. Fig. \ref{fig: large_dialog} illustrates examples that the \modelname\ takes actions under different user feedback settings.

{\bf Baselines.} Three strong zero-shot object navigation baselines are adapted to our flexible \modelname\ framework. 
% We compare our method against three strong baselines models, which demonstrate good capabilities in zero-shot language-based navigation:
\begin{itemize}
    \item CLIP-on-Wheels (CoW) \cite{cow} uses FBE to explore and localize objects in images with CLIP saliency maps using Grad-CAM \cite{selvaraju2017grad}. 
    We apply these to the exploration module and open-vocabulary detection module. However, CoW does not have a semantic map and memory module originally.
    \item VLMap \cite{vlmap} builds a spatial map representation that directly fuses pretrained visual-language features (e.g., Lseg) with a 3D reconstruction of the physical world. The map enables natural language indexing for zero-shot navigation using CLIP text features. We apply it to the semantic map module, but VLMap does not contain detection, memory and exploration modules.
    \item ConceptFusion (CF) \cite{jatavallabhula2023conceptfusion} adopts a similar map creation technique but maintains original CLIP vision  features to capture uncommon objects with higher recall than VLmap. We utilize the CF map in the semantic map module, while other modules remain the same as VLMap.
\end{itemize}
All compared methods are connected with the same LLM (GPT-4-8k-0613) to schedule different modules to navigate and interact with users.

\begin{table*}[t]
\centering
\scalebox{0.9}{
\begin{tabular}{lcccccccccccccccccc}
\toprule
\multirow{2}{*}{Method} & \multicolumn{3}{c|}{No Interaction} & \multicolumn{3}{c|}{Yes/no Feedback} & \multicolumn{3}{c|}{Corrective Feedback} & \multicolumn{3}{c|}{Descriptive Feedback} & \multicolumn{3}{c|}{Landmark Feedback} & \multicolumn{3}{c}{Procedure Feedback} \\  
 & \multicolumn{1}{c}{SR} & \multicolumn{1}{c}{SPL} & \multicolumn{1}{c|}{SIT}
 & \multicolumn{1}{c}{SR} & \multicolumn{1}{c}{SPL} & \multicolumn{1}{c|}{SIT} & \multicolumn{1}{c}{SR} & \multicolumn{1}{c}{SPL} & \multicolumn{1}{c|}{SIT} & \multicolumn{1}{c}{SR} & \multicolumn{1}{c}{SPL} & \multicolumn{1}{c|}{SIT} & \multicolumn{1}{c}{SR} & \multicolumn{1}{c}{SPL} & \multicolumn{1}{c|}{SIT} & \multicolumn{1}{c}{SR} & \multicolumn{1}{c}{SPL} & \multicolumn{1}{c}{SIT} \\ \midrule
Human & -- & -- & -- &  -- & -- & -- & 94.5 & 75.7 & 81.9 & 94.5 & 75.4 & 84.8 & 95.0 & 76.6 & 86.0 & 97.2 & 78.9 & 86.9\\
\midrule
COW & 15.4 & 8.4 & 15.4 & 36.8 & 24.2 & 35.8 & 38.5 & 21.6 & 29.0 & 53.5 & 22.4 & 33.2 & 43.9 & 21.3 & 30.2 & 59.0 & 22.5 & 32.8 \\
VLmap & 23.9 & 20.3 & 23.9 & 41.8 & 30.8 & 31.2 & 43.6 & 32.6 & 35.2 & 44.4 & 31.5 & 35.8 & 53.3 & 35.8 & 39.0 & 67.5 & \textbf{52.7} & 44.1 \\
CF &21.3 & 13.7 & 21.3 & 47.9 & 25.6 & 31.1 & 52.1 & \textbf{37.5} & 36.5 & 47.0 & 33.5 & 34.1 & 59.9 & \textbf{40.1} & 40.6 & 68.4 & 49.4 & 39.3\\
ORION & \textbf{28.2} & \textbf{24.9} & \textbf{28.2} & \textbf{54.2}$^*$ & \textbf{35.5} & \textbf{37.8} & \textbf{59.0}$^*$ & 34.2 & \textbf{39.1} & \textbf{63.7}$^*$ & \textbf{37.4} & \textbf{44.2}$^*$ & \textbf{69.5}$^*$ & \textbf{40.1} & \textbf{46.8} & \textbf{80.3}$^*$ & 51.1 & \textbf{52.8}$^*$ \\ \bottomrule
\end{tabular}
}
\caption{: Results of different methods in single-feedback settings with simulated users. The asterisk * indicates a statistically significant improvement of \modelname\ over all three baselines for each column (wilcoxon test; p $<$ 0.05).}
\vspace{-14pt}
\label{tab: main_sim}
\end{table*}

\begin{table}[t]
\centering
\scalebox{0.9}{
\begin{tabular}{lrrr}
\toprule
Method & \multicolumn{1}{c}{SR} & \multicolumn{1}{c}{SPL} & \multicolumn{1}{c}{SIT} \\ \midrule
% Human & 96.3 & 79.7 & 88.0 \\
% \midrule
COW & 62.4	& 33.7	& 36.2\\
VLMap & 71.8 & 54.8 & 48.0\\
CF &73.5&	52.3&	44.2\\
ORION & \textbf{83.8$^*$} & \textbf{56.6} & \textbf{53.5} \\
\midrule
\ \ \  w/o mem & 81.2 & 54.8 & 51.2 \\
\ \ \  w/o exp & 75.9 & 51.6 & 49.7 \\ 
\ \ \   w/o det & 72.5 & 47.7 & 42.2 \\
\ \ \   w/o map & 69.2 & 37.5 & 41.3 \\
\bottomrule`
\end{tabular}}
\vspace{-10pt}
\caption{: \small Results of different methods (upper part) and ablations (lower part) in the mixed-feedback setting with simulated users.
% * means statistically significance as in Tab. \ref{tab: main_sim}
}
\vspace{-20pt}
\label{tab: mixed-ablation}
\end{table}

{\bf Evaluation Metrics.} A good interactive navigation agent should be efficient (i.e., achieve the goal as fast as possible) and pleasant (i.e., bother the user as little as possible). Therefore, 
we use (i) Success Rate (SR), defined as $\text{SR}=\frac{1}{N}\sum_{i=1}^NS_i$ to evaluate the percentage that the agent reaches the correct goals, $S_i \in \{0,1\}$ is the binary indicator of success for $g_i$, $N$ is the total number of goals; 
(ii) Success Rate weighted by the Path Length (SPL) \cite{habitatchallenge2022}, defined as $\text{SPL}=\frac{1}{N}\sum_{i=1}^NS_i\frac{l_i}{max(a_i, l_i)}$, where the $l_i$ denotes the ground truth shortest path length, and $a_i$ denotes the actual path length;
(iii) Success Rate weighted by the Interaction Turns (SIT), defined as $\text{SIT}=\frac{1}{N}\sum_{i=1}^NS_i\frac{1}{I_i}$, where $I_i \geq 1 $ is the number of interactions between the agent and user for $g_i$. SPL and SIT indicate the navigation and interaction efficiency respectively.

\subsection{Evaluation with Simulated Environment}
In the simulated environment, we build simulated users to scale up the experiments and compare different methods under various feedback settings.

\label{sec: exp_sim}
\subsubsection{Experimental Setup} We use the Habitat simulator \cite{savva2019habitat} and the high-quality realistic indoor scene dataset HM3D v0.2 \cite{ramakrishnan2021habitat} for 
experiments. Ten scenes are randomly selected from the validation set. A total of 14,159 RGB-D frames are collected for semantic map creation, and 117 goal objects are randomly selected for evaluation. To construct the personalized goals, we annotate the chosen objects with various types of personal information, such as people’s names (e.g., \textit{Alice’s computer}), manufacturers of products (e.g., \textit{chair from IKEA}) and purchase dates (e.g., \textit{bed bought last year}), so that each $g$ in $\mathcal{G}$ can be uniquely identified. We manually write object descriptions for each $g$ for the descriptive feedback. For the procedural feedback, we generate the
ground-truth geodesic path and translate it into language sentences to indicate an approximate route, e.g., \textit{``turn left,
go forward 5 meters, turn right”}.
% Fig. X illustrates the personalized goal curation. 
The fast-marching method \cite{sethian1996fast} is used for low-level motion planning. 
The map size $L$x$W$ is 600x600, where each grid equals 0.05m in the environment. The RGB-D image frame is 480x640 with a camera fov 90 degrees. The depth range is set in [0.1m,10m] for map processing. CLIP-ViT-B-32 is used for text feature extraction and semantic matching for the semantic map and memory module. 
An episode is successful when the agent is within 1.5 meters and the mass centre of $g$ is in the ego-view image. $I_{\max}$ is 5 for each goal.

% Given a test scene, we first choose a set of common objects like ``bed" as the naive goals from the ground-truth object set, then we add personalized information and diverse language feedback to extend into personalized goals.

% {\bf Personalized Goals.} 
% We first randomly select a set of people's names and assign different names to multiple same-typed rooms in the scene, e.g., if there are two bedrooms, we then randomly name them as ``Alice's bedroom" and ``Bob's bedroom" from the name list; if only one living room, the room name remains ``living room".  After a room belongs to a person, then by default all the goal objects in the room belonging to that person so that the user can issue commands like ``find Alice's computer" to the robot. We also generate a set of product information, such as company names and object materials and assign them to different goals, e.g., ``chair bought from IKEA" and ``table made of mahogany", so that each $g$ in $\mathcal{G}$ can be uniquely identified.

\subsubsection{User Simulator} As real user interactions can be tedious and time-consuming, simulated users are often used as a substitute to evaluate dialogue agents \cite{keizer2010parameter, kreyssig-etal-2018-neural,tseng-etal-2021-transferable}. We use another LLM (dubbed as user-LLM) as the backbone to build a user simulator to interact with robots. At each turn, all relevant ground-truth information is sent as input to the user-LLM in a dictionary, which includes the dialogue context, personalized goal information, task success signals, robot detection results, etc. Then, the user-LLM generates suitable user utterances for the robot, guided by appropriate prompt examples. 
We chose GPT-3.5-turbo-0613 since we found it adequate to produce reasonable user utterances. Each scene is run 5 times with random seeds to get average results.

\begin{figure}[t]
    \centering
    \includegraphics[width=0.8\linewidth]{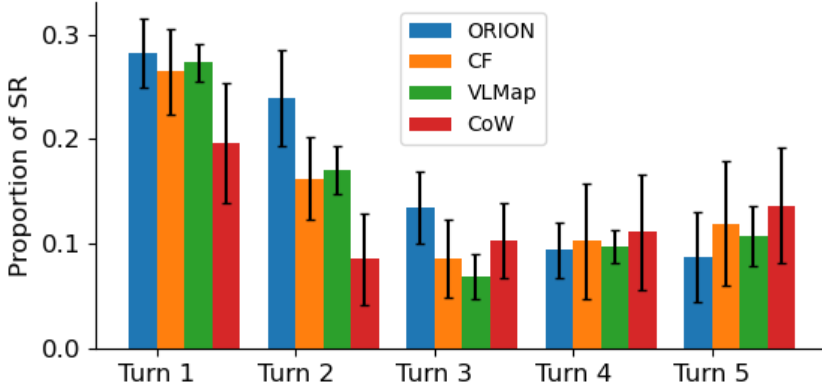}
    \caption{\small Distribution of task success rates based on interaction turns (turn 1-5) in the mixed feedback setting for all compared methods.}
    \label{fig: distribution}
    \vspace{-16pt}
\end{figure}

\subsubsection{Evaluation Results}
\label{sec: main_sim}
Tab. \ref{tab: main_sim} shows the results of all methods under single-feedback settings, where we control the input with only one type of feedback to be sent to the user-LLM to generate the user utterance.
For comparison, we add \text{\small  `No Interaction'} setting, which means the robot can not interact with users, and 
\text{\small  `Yes/no Feedback'} setting, which means the user only gives yes/no response as minimal information to indicate whether the agent has reached the goal without any extra feedback information. We also include the results of human-teleoperated agents as the upper bound. 

{\bf Comparison on different types of feedback.} 
As shown in Tab. \ref{tab: main_sim}, using each type of feedback contributes to a performance increase to some degree, highlighting the significance of natural language interactions. Specifically, the procedure feedback brings the most significant improvement, with an increase of 21-26\% SR across all methods compared to the Yes/no Feedback. We found this because it can suggest a route or a destination point even though it is not precise. This guidance helps agents transition to nearby areas,  substantially narrowing the scope needed to locate the goal. The landmark feedback also largely boosts the performance 
% especially for the map-based methods, e.g., 10\% improvements for VLMap and CF, and 13\% improvements for ORION. This is because \texttt{retrieve\_map} can return all possible matched nearby objects in the semantic map module, so the agent can issue \texttt{goto\_object} to go directly to the retrieved nearby object for further goal searching. This feedback is also important for exploration-based methods like CoW 
since the provided nearby objects could be easier to find than the goal objects, thus helping the robot reach the goal's neighbourhood. 
Comparatively, the descriptive feedback does not yield much benefit but still boosts methods that have the exploration module, resulting in a 16.7\% and 10.5\% increase in SR for CoW and ORION, respectively.
We conjecture this is because those methods use open-vocabulary detection models to process online ego-view RGB images, so the fine-grained visual semantics of the objects can be maintained to align with rich language descriptions, whereas map-based methods like VLMap may lose these nuances during map creation. The corrective feedback brings the least improvement since it only improves when the wrong objects the agent found for the current goal could happen to be some correct objects of other goals in the same scene. 

Besides the single-feedback settings
% , where only one type of feedback information was utilized throughout the evaluation, 
we also evaluated in the mixed-feedback setting, where the user-LLM receives all types of feedback information to generate utterances.
% \jycc{user simulator create all these different types at the same time? if not, again the same question, why do they know which feedback to choose, randomly? } \dyp{yes, it creates all at the same time with various language expressions. This is in-turn mixed feedback.}
As observed in the upper section of Tab. \ref{tab: mixed-ablation}, all methods can be future improved from single-feedback settings, showing approximately 3-5\% in both SR and SIT. This indicates that the richer the information a robot can access during interaction, the better its performance in the environment.

{\bf Comparison on different methods}
% From the results in Tab \ref{tab: main_sim} and \ref{tab: mixed-ablation}, we can see that no single method can significantly dominate all metrics in \taskname\ tasks. In this section, we discuss in detail the current challenges that different methods encounter.  (To be tighten up..)
From Tabs \ref{tab: main_sim} and \ref{tab: mixed-ablation}, it's evident that no single method consistently outperforms across all metrics in \taskname\ tasks. Key challenges include:
(i) \textit{Balancing task success with navigation efficiency}. While \modelname\ can achieve a high SR, it often lags in SPL, e.g., in Tab. \ref{tab: main_sim}, it surpasses VLMap in SR by over 10\% but is only 2\% ahead in SPL. This is because it can explore and actively move in the room to detect, often involving more steps to find the goal. But map-based methods take direct movements to retrieved objects, regardless of their accuracy.
% This will completely rely on the semantic map quality, thus leading to limited SR.
% Even though our \modelname\ combines both the semantic map module and the exploration module, the LLM tends to use detection actions like \texttt{double\_check} to move around to double-check the ego-view images or start to \texttt{seach\_object} to search the whole room if the semantic map provides wrong objects for several turns.
% But for the SIT metrics, our \modelname\  outperforms on all setting, suggesting it is more suitable for interactive navigation with users.
(ii) \textit{Balancing task success with interaction efficiency}. Compared with human tele-operated agents, all methods suffer from low SIT performance, 
% For our \modelname, even though it achieves the highest success rate (83.8\%) in the mixed-feedback setting, the SIT is still 53.5\%, 
indicating a large dependence on interaction rather than finding the goal objects more efficiently. 
% In contrast, the human tele-operated agent can obtain 88\% SIT. 
Fig \ref{fig: distribution} further illustrates this, showing that a considerable proportion of SR requires large interaction turns for all methods. 
{\bf Ablations and more analysis.}
\label{sec: abl}
% The lower part of Tab. \ref{tab: mixed-ablation} gives the ablation results on different modules for \modelname\ under the mixed feedback setting. In the table, {w/o mem} is the \modelname\  without using the additional memory to save the information;  {w/o exp} means we do not use the exploration module, but we still allow the agent to perform 360$^\circ$ spinning when calling \texttt{search\_object} as 
% a restricted exploration strategy; {w/o det} is the \modelname\  that does not use the grounded-SAM model, we replace it with a simple k-patch mechanism \cite{cow} that CLIP is used to match the text descriptions and each of the 3x3 patches in the image; {w/o map} is the \modelname\  without the semantic map module. 
Tab. \ref{tab: mixed-ablation}'s lower section presents ablation results for \modelname\ in the mixed feedback setting. Here, \text{\small `w/o mem'} excludes the additional memory feature;  \text{\small `w/o exp'} omits the exploration module but retains the 360$^\circ$ spin;  \text{\small `w/o det'} replaces the grounded-SAM model with a basic k-patch mechanism \cite{cow} where the CLIP matches texts to image patches; and  \text{\small `w/o map'} removes the semantic map module.
The results emphasize the crucial role of the semantic map, marked by the most significant SR drop in \text{\small \modelname\ w/o map}. 
Both \text{\small  \modelname\ w/o exp} and \text{\small \modelname\ w/o det} yield comparable outcomes, suggesting the importance of a robust detection model paired with active exploration. Interestingly, the memory module has a limited impact, even with its capability to retain user information. We hypothesize this is because, during the zero-shot evaluation, previously stored goals aren't retested. To further explore this, we conduct a second time 
\taskname\ evaluation using the memory accumulated from the first time. Consequently, \modelname\ gets scores of 91.5\% in SR, 63.9\% in SPL, and 65.3\% in SIT.

\subsection{Evaluation with Real Robots}
\label{sec: realrobot}
We also perform real-world experiments with the TIAGo robot for the indoor \taskname\ using the \modelname\ framework.
\subsubsection{Experimental Setup} We select 20 goal objects in a room for navigation. 
% including 12 large objects like a table and 8 small objects like a water bottle.
Each object is assigned a unique person's name unless it's paired with another object. For instance, Alice's computer would be on Alice's table. We then manually provide 3-5 sentences for each goal as the descriptive feedback. Nine salient objects (e.g., fridge) in the room are used as landmarks. Simple instructions like \textit{``3 meters to your left"} are used for the procedural feedback. The built-in GMapping \cite{gmapping} and move\_base package are used for SLAM and path planning.
% and move\_base in the ROS navigation stack for path planning and robot moving.
To create the semantic map, we construct a topology graph that includes the generic classes of all landmark objects and large goal objects for simplicity. Experiments are run twice to get average results. Other set-ups remain the same with simulated experiments.
% To enhance exploration speed, we set 6 pre-defined viewpoints in the room for the robot to access. 
% Other set-ups remain the same with simulated experiments.
% consistent with those in the simulated environment.

\begin{table}[h]
\centering
\vspace{-5pt}
\scalebox{0.9}{
\begin{tabular}{cccc}
\toprule
Feedback Type & SR & SPL & SIT \\ \midrule
No Interaction & 37.5 & 35.8 & 37.5 \\
\midrule
% Yes/no & 50.0 & 37.7 & 35.8 \\
Corrective & 67.5$^*$ & 51.8$^*$ & 53.6$^*$ \\
Descriptive & 67.5$^*$ & 48.6 & 48.0 \\
Landmark & \textbf{72.5$^*$} & \textbf{66.8$^*$} & 54.8$^*$ \\
Procedural & \textbf{72.5$^*$} & 62.8$^*$ & \textbf{61.9$^*$} \\
% Mixed & 55.0 & 53.3 & 42.9 \\
\bottomrule
\end{tabular}}
\caption{: Results on real robots. $^*$ means statistically significant difference compared to \text{\small `No Interaction'} (wilcoxon test; p$<$0.05)}
\label{tab: real_robot}
\vspace{-10pt}
\end{table}

\subsubsection{Evaluation Results}
Tab. \ref{tab: real_robot} displays the results of \modelname\ under different feedback settings. We can see that leveraging user feedback enhances overall performance with similar trends in simulated environment results.
Specifically, the landmark feedback yields substantial improvement in SR and SPL, as the selected landmarks are easily identifiable in the room, thus effectively narrowing down the search.
While the procedural feedback provides only basic navigational cues about the goal, it still greatly enhances SIT.
Given the room's straightforward layout, the robot often encounters potential goals during evaluation; therefore, the corrective feedback also largely improves the results by rectifying misidentified objects for  \modelname\ to update its memory.
Comparatively, the descriptive feedback brings the least SPL and SIT as matching real-world observations with language descriptions using VLMs is still quite challenging. 
Common failure cases stem from two main sources: 1) low-level movement errors, which negatively impact task success performance; and 2) inaccurate detection, which fails to identify the correct objects in the robot's ego-view images.
% To explore the robot performance interacted with real users, we ask xxxx users to communicate with the robot using the same goals but allow them to say any free-formed feedback. In the end,  \modelname\ gets xxx in SR, xxx in SPL and xXX in SIT, indicating xxxx. 
\section{Conclusion}
This work introduces Zero-shot Interactive Personalized Object Navigation (\taskname), an advanced version of zero-shot object navigation. In this task, a robot needs to navigate to personalized goal objects while engaging in natural language interactions with the user. To address the problem, we propose \modelname, a general framework for open-world interactive personalized navigation, where the LLM serves as a decision-maker to direct different modules to search, perceive, navigate in the environment, and interact with the user. 
Our results in both simulated environments and the real world demonstrate the utilities of different types of language feedback. They also point out the challenges to obtaining a good balance between task success, navigation efficiency, and interaction efficiency. These findings will provide insights for future work on language communication for human-robot collaboration. 
This work is only our initial step in exploring LLMs in personalized navigation and has several limitations. For example, it does not handle broader goal types, such as image goals,  or address multi-modal interactions with users in the real world. Our future efforts will expand on these dimensions to advance the adaptability and versatility of interactive robots in the human world. 

% In the future, we will extend this work to more general settings and solve real problems. We hope the proposed task and framework will explore broader and more diverse forms of zero-shot embodied AI.


%%%%%%%%%%%%%%%%%%%%%%%%%%%%%%%%%%%%%%%%%%%%%%%%%%%%%%%%%%%%%%%%%%%%%%%%%%%%%%%%

%%%%%%%%%%%%%%%%%%%%%%%%%%%%%%%%%%%%%%%%%%%%%%%%%%%%%%%%%%%%%%%%%%%%%%%%%%%%%%%%

%%%%%%%%%%%%%%%%%%%%%%%%%%%%%%%%%%%%%%%%%%%%%%%%%%%%%%%%%%%%%%%%%%%%%%%%%%%%%%%%

\clearpage

\bibliographystyle{unsrt}
\bibliography{citation}

% \input{appendix}


% \section*{ACKNOWLEDGMENT}

% The preferred spelling of the word ÒacknowledgmentÓ in America is without an ÒeÓ after the ÒgÓ. Avoid the stilted expression, ÒOne of us (R. B. G.) thanks . . .Ó  Instead, try ÒR. B. G. thanksÓ. Put sponsor acknowledgments in the unnumbered footnote on the first page.

%%%%%%%%%%%%%%%%%%%%%%%%%%%%%%%%%%%%%%%%%%%%%%%%%%%%%%%%%%%%%%%%%%%%%%%%%%%%%%%%

\end{document}